\documentclass{article} 
\usepackage{iclr2015,times}
\usepackage{hyperref}
\usepackage{url}
\usepackage{amsmath}
\usepackage{amssymb}
\usepackage{multirow}
\usepackage{amsthm}
\usepackage{algorithm}
\usepackage{algorithmic}
\newtheorem{mydef}{Definition}
\newtheorem{mycor}{Corollary}
\def\etal{\emph{et al }}
\iclrconference
\usepackage{graphicx}

\title{Cauchy Principal Component Analysis}

\author{
Pengtao Xie \& Eric Xing \\
School of Computer Science \\
Carnegie Mellon University \\
Pittsburgh, PA, 15213 \\
\texttt{\{pengtaox,epxing\}@cs.cmu.edu} 
}

\begin{document}

\maketitle

\begin{abstract}
  Principal Component Analysis (PCA) has wide applications in machine learning, text mining and computer vision.
Classical PCA based on a Gaussian noise model is fragile to noise of large magnitude.
Laplace noise assumption based PCA methods cannot deal with dense noise effectively.
In this paper, we propose Cauchy Principal Component Analysis (Cauchy PCA), a very simple yet effective
  PCA method which is robust to various types of noise. 
  We utilize Cauchy distribution to model noise and derive Cauchy PCA under the maximum likelihood estimation (MLE) framework with low rank constraint. Our method can robustly estimate the low rank matrix regardless of whether
  noise is large or small, dense or sparse. 
  We analyze the robustness of Cauchy PCA from a robust statistics view and present an efficient singular value projection optimization method.
  Experimental results on both simulated data and real applications demonstrate
  the robustness of Cauchy PCA to various noise patterns.
\end{abstract}

\section{Introduction}

Principal Component Analysis and related subspace learning techniques based on matrix factorization
have been widely used in 
dimensionality reduction, data compression, image processing, feature extraction and data visualization.
It is well known that PCA based on a Gaussian noise model (i.e., Gaussian PCA) is sensitive to noise of large magnitude, because the effects of large noise are exaggerated by the use of Gaussian distribution induced quadratic loss. 
To robustify PCA, a number of improvements have been proposed \citep{de2003framework,khan2004robust,ke2005robust,archambeau2006robust,ding2006r,brubaker2009robust,candes2009robust, eriksson2010efficient}. 
Roughly, these methods can be grouped into a non-probabilistic paradigm and a probabilistic paradigm according to whether they are built on a probabilistic assumption of the noise. The non-probabilistic approaches \citep{de2003framework,ding2006r,brubaker2009robust} either use robust $p$-function to weight the squared loss of each data item according to its fitness to the subspace, or try to robustly estimate the covariance matrix by alternatively removing or down-weighting samples corrupted with large noise. Such a non-probabilistic paradigm makes it difficult, if possible, to take advantage of some desirable utilities offered by sophisticated probabilistic models such as Bayesian treatment of PCA, mixture of probabilistic PCAs \citep{tipping1999probabilistic} and probabilistic matrix factorization \citep{salakhutdinov2008probabilistic},
and do not facilitate statistical testing or comparison with other probabilistic techniques.


The probabilistic robust PCA methods \citep{khan2004robust,ke2005robust,archambeau2006robust,candes2009robust,eriksson2010efficient} are derived by replacing the Gaussian assumption of the noise with a Laplace assumption \citep{ke2005robust,candes2009robust,eriksson2010efficient} or Student-t assumption \citep{khan2004robust,archambeau2006robust}. Both Laplace distribution and Student-t distribution have heavy tails which can reasonably explain data far away from the mean, thus they are suitable for modeling spiky noises with large magnitude. However, these methods suffer new problems.
A major drawback of Laplace PCA is that it is incapable of coping with dense noise. This is because a Laplace distribution and the resultant $\ell_{1}$ norm would induce sparsity in the solution \citep{tibshirani1996regression}, thereby falsely using a sparse model to explain dense noise. 
Student-t distribution can avoid the drawbacks of Laplace distribution and Gaussian distribution. \citet{khan2004robust} and \citet{archambeau2006robust} tried to robustify probabilistic PCA (PPCA) \citep{tipping1999probabilistic} by replacing the Gaussian noise assumption with Student-t assumption.
In practice, we find their methods have very similar performance with Gaussian PCA and work much worse on large noises than Laplace PCA .

In our opinion, data noises can be roughly partitioned into four patterns according to their abundance and magnitude: sparse small noise, sparse large noise, dense small noise and dense large noise. Gaussian PCA is limited to small noise and Laplace PCA is only suitable for sparse noise. Once noise is both large and dense, neither of the two PCA methods will be suffice.
In reality, dense large noise is quite ubiquitous. For instance, in factorization based structure from motion \citep{tomasi1992shape}, due to bad illumination, fast optical flow, occlusions, and the deficiency of tracking algorithms, grossly mistracked features are quite common.
In photo sharing websites (like Flickr, Instagram), many user generated tags are irrelevant to the images and many objects and attributes in images are not tagged by users. Thus considerable false positives and false negatives are present in the data.
With the popularity of low cost cameras on mobile devices, millions of user generated videos are published to video sharing websites like Youtube. Due to high capturing rate, poor light conditions and users' unprofessional capturing habits, videos are usually contaminated with gross noise affecting nearly every pixel, especially when videos are taken at night or on fast moving vehicles. On the other hand, in many problems, noise patterns are mixed. It is common that most entries of the low rank matrix are corrupted by small noise while a small part are contaminated by large noise. Gaussian PCA and Laplace PCA are not applicable in this case since neither of them are able to deal with the two types of noise simultaneously. \citet{zhou2010stable} proposed Stable Principal Component Pursuit to recover matrices corrupted by small entry-wise noise and gross sparse errors. However, their method requires a good estimation of the magnitude of small noise, which is infeasible in many real applications.

In this paper, we propose an alternative probabilistic robust PCA method called Cauchy PCA, which is robust to all kinds of noise patters. We use Cauchy distribution to model noise and derive Cauchy PCA under a maximum likelihood estimation framework with rank constraints. We present a simple yet efficient projected gradient optimization method. Experiments demonstrate the robustness of Cauchy PCA to various noise patterns,
and in particular, its superior capability in dealing with large dense noise.

The rest of the paper is organized as follows. Section 2 introduces related work. We propose Cauchy PCA in section 3. Section 4 gives experimental results on both simulated data and real world data. Section 5 concludes the paper.

\section{Related Works}
Robust PCA methods can be categorized into two paradigms: non-probabilistic approaches and probabilistic approaches. The basic strategy of non-probabilistic methods is to remove or down-weight the influence of large noise corrupted data items. \citet{de2003framework} proposed robust subspace learning by replacing the squared loss function in PCA with Geman-McClure error function which is less sensitive to large noise and used iteratively reweighted least squares (IRLS) method to solve the problem. \citet{ding2006r} proposed a rotational invariant $\ell_{1}$ norm PCA, whose principal components are eigenvectors of a re-weighted covariance matrix which softens samples corrupted with large noise. \citet{brubaker2009robust} estimated the subspace by alternatively removing outliers and projecting to a lower dimensional subspace. These methods lack the ability to build or integrate with sophisticated probabilistic models.

In probabilistic approaches, one popular family is based on Laplace noise assumption and $\ell_{l}$ norm.
\citet{ke2005robust} proposed a matrix factorization formulation with $\ell_{l}$ norm and used alternating convex optimization as a solver. \citet{candes2009robust} proposed Principal Component Pursuit (PCP) to recover the low rank matrix corrupted by sparse errors of arbitrarily large magnitude. They prove that, as if the noise is sufficiently sparse and the rank of the true underlying subspace is sufficiently small, the low rank matrix can be exactly recovered. \citet{eriksson2010efficient} generalized the Wiberg algorithm to solve the $\ell_{1}$ norm based low rank matrix approximation problem in the presence of missing data. All these $\ell_{1}$ norm (Laplace noise assumption) based methods are incapable of dealing with dense noise. \citet{ganesh2010dense} claimed that by choosing a proper value of the tradeoff parameter, Principal Component Pursuit is also robust to dense noise. However, we find in practice, their suggested way of choosing parameter yields very poor results. 
\citet{zhou2010stable} incorporated an additional term into the PCP model to account for small dense noise, but it is still unable to handle large dense noise. 
\citet{xu2010robust} formulated a similar problem as \citet{candes2009robust} to identify entirely corrupted points by imposing $\ell_{1,2}$ norm on the noise matrix. Their method assumes column level corruption of the low rank matrix and is not applicable for entry level corruption.

Another family of probabilistic PCA methods \citep{archambeau2006robust,khan2004robust} are based on Student-t distribution. 
They assume observed data $\mathbf{y} \in \mathbb{R}^{D}$ are generated as $\mathbf{y}=\mathbf{W}\mathbf{x}+\mathbf{u}+\mathbf{n}$, where $\mathbf{x}\in \mathbb{R}^{d}$ is a latent vector in a lower-dimensional space, $\mathbf{W}\in \mathbb{R}^{D\times d}$ is the projection matrix, $\mathbf{u}\in \mathbb{R}^{D}$ is data offset and $\mathbf{n}\in \mathbb{R}^{D}$ is noise drawn from Student-t distribution.
They utilize the property that Student-t distribution is a infinite mixture of Gaussian distributions with the same mean and varying variations and propose an expectation maximization (EM) algorithm to infer latent vectors $\mathbf{x}$ and learn parameters $\mathbf{W}$ and $\mathbf{u}$. Empirically, these methods are comparable with Gaussian PCA on small noise. For large noise, they work slightly better than Gaussian PCA but much worse than Laplace PCA.

\section{Cauchy Principal Component Analysis}
In this section, we first present location-scale family distributions and shows how they can be used to derive PCA methods. Then we present some intuition of choosing Cauchy distribution to model noise by comparing its density curve with other distributions in location-scale family. We introduce Cauchy PCA by specializing the general location-scale family PCA framework with Cauchy distribution, interpret its robustness from a robust statistics view and propose an efficient singular value projection solver.
%

\begin{figure}[t]
\begin{center}
   \includegraphics[width=\linewidth]{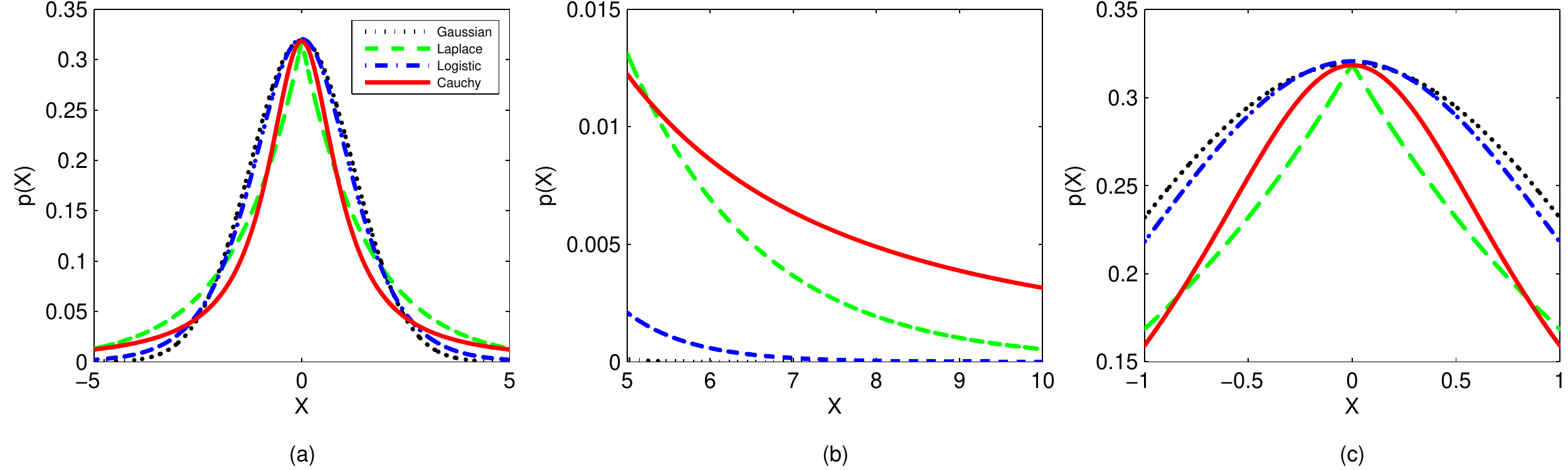}
\end{center}
   \caption{Density curves of Gaussian, Laplace, Logistic and Cauchy distribution over different ranges of random variable $X$. All curves are aligned to be peaked at zero and share the same peak value.}
\label{density_curves}
\end{figure}

\subsection{Location-Scale Family PCA}\label{sec:lsf}
Location-scale family is a class of distributions parameterized by a location parameter and a scale parameter. The most important property of this family is that distributions are closed under linear transform. If $X$ is a random variable drawn from this family, then $aX+b$ is also from this family. This property provides convenience to model additive noise. In PCA setting, assume each entry of noise matrix $E$ is from i.i.d location-scale family distribution
\begin{equation} \label{eq:lcdis}
E_{ij}\sim p(E_{ij}|0,s)
\end{equation}
 with location parameter zero and scale parameter $s$. According to the closure under linear transformation property and additive noise assumption $M=L+E$, observation matrix $M$ can be modeled as
 \begin{equation}
 M_{ij} \sim p(M_{ij}|L_{ij},s)
 \end{equation}
 with shifted location parameter $L_{ij}$. $L$ can be estimated by maximizing the likelihood of observations (or minimizing the negative log likelihood) with low rank constraint
\begin{equation}
\begin{array}{ll}
\text{max}_{L} & \prod_{ij}p(M_{ij}|L_{ij},s)\\
\text{s.t.}& rank(L)\leq k\\
\end{array}
\end{equation}
Gaussian PCA and Laplace PCA \citep{candes2009robust,ke2005robust} are special cases of the general framework by specifying the distribution in Eq.(\ref{eq:lcdis}) to Gaussian and Laplace distribution respectively. 
\subsection{Cauchy PCA}\label{sec:cpca}
Figure \ref{density_curves} shows the density curves of univariate Gaussian, Laplace, Logistic and Cauchy distributions. 
To enable a clear comparison, density curves are aligned to the same location and peak. The motivation of aligning their peaks is to inspect an interesting phenomenon: if we put the same amount of probability on the mode of each distribution, how much probability will each distribution allocates for other values? This can give us a good sense of heavy-tail-ness.
 As data points get far away from the center, Gaussian probability drops quickly to zero while Laplace and Cauchy probabilities remain a certain amount as shown in Figure \ref{density_curves}(b). In other words, Laplace and Cauchy density curves have longer tails than Gaussian curve. A distribution (centered at zero) with heavy tail allocates a reasonable amount of probability on values far from zero. In terms of noise modeling under a probabilistic framework, large noises can be reasonably explained by heavy tail distribution since a certain amount of probability is granted to them. Thereby, Laplace PCA and Cauchy PCA naturally possess the ability of dealing with large noise due to their heavy-tail-ness. At location zero (Figure \ref{density_curves}(c)), Laplace distribution is not differentiable. The non-smoothness property induces sparsity, which makes Laplace distribution unsuitable to model dense noise. Logistic distribution highly resembles Gaussian distribution in shape except a slightly heavier tail. Therefore its behavior in modeling noise should be very similar to Gaussian distribution. Among the four, Cauchy distribution owns two appealing advantages. First, it is smooth at zero and does not induce sparsity, thus is suitable for modeling dense noise. Second, it has a much heavier tail than the others, therefore, it is highly capable of modeling large noise.

We use Cauchy distribution with location parameter zero to model noise $E$
\begin{equation}
p(E_{ij})=\frac{\gamma}{\pi}\frac{1}{\gamma^{2}+E_{ij}^2}
\end{equation}
where $\gamma$ is the scale parameter. Substituting it into Eq.(\ref{eq:lcdis}), we specialize the general location-scale family PCA framework to Cauchy PCA
\begin{equation}\label{eq:cpca}
\begin{array}{ll}
\text{min}_{L} & \sum_{ij}\log(\gamma^{2}+(M_{ij}-L_{ij})^2)\\
\text{s.t.}& rank(L)\leq k\\
\end{array}
\end{equation}

Cauchy PCA can be naturally extended to deal with missing data. We use $I_{ij}=1$ to denote that the entry at the $i$th row and $j$th column of $M$ is observed, $I_{ij}=0$ otherwise. We maximize the following data likelihood
\begin{equation}
\begin{array}{ll}
\text{max}_{L} & \prod_{ij}p(M_{ij}|L_{ij},s)^{I_{ij}}\\
\text{s.t.}& rank(L)\leq k\\
\end{array}
\end{equation}
which is equivalent to introducing a 0-1 weight matrix to weight each item in Eq.(\ref{eq:cpca}).


\subsection{A Robust Statistics Interpretation}
In this section, we explain the robustness of Cauchy PCA from a robust statistics view \citep{hampel2011robust}. Robust statistics seek to provide robust estimators resisting against gross noise. To be consistent with Section \ref{sec:lsf} and \ref{sec:cpca}, we assume distributions are located at zero and parameters are estimated using maximum likelihood estimation (MLE).
For a set of distributions $F$ and the estimator $T$ defined on $F$, Hample \etal~ (1974) introduced the influence function:
\begin{mydef}
The influence function (IF) of $T$ at $F$ is given by
\begin{equation}
\text{IF}(x;T,F) = \lim_{t \to 0}\frac{T((1-t)F+t\Delta_{x})-T(F)}{t}
\end{equation}
in those $x \in \mathbb{R}$ where this limit exists.
\end{mydef}
Heuristically, influence function describes the effect of an infinitesimal contamination at the point $x$ on the estimation.
Based on IF, Hampel \etal~ (1974) defined \textit{gross-error sensitivity}:
\begin{mydef}
The gross-error sensitivity of $T$ at $F$ is measured by
\begin{equation}
\gamma^{*}=\sup_{x} |\text{IF}(x;T,F)|
\end{equation}
the supremum being taken over all $x$ where IF exists.
\end{mydef}
The gross-error sensitivity measures the worst influence which a small amount of contamination of fixed size can have on the estimator. A desirable robust estimator should have finite $\gamma^{*}$.
\begin{mycor} \label{coro::ges}
$\gamma^{*}$ of Cauchy and Laplace MLE estimators are bounded. $\gamma^{*}$ of Gaussian MLE estimator is unbounded.\footnote{Due to space limit, the proof of Corollary 1 and 2 are provided in supplementary meterial.}
\end{mycor}
Corollary \ref{coro::ges} explains why Cauchy and Laplace PCA are robust to gross noise while Gaussian PCA is not.

Another quantity \textit{local-shift sensitivity} \citep{hampel1974influence} is defined to measure the effect on estimators by shifting an observation slightly from one point to some neighboring point:
\begin{mydef}
The local-shift sensitivity of an estimator $T$ is defined as
\begin{equation}
\lambda^{*}=\sup_{x\neq y}|\text{IF}(y;T,F)-\text{IF}(x;T,F)|/|y-x|
\end{equation}
\end{mydef}
A stable and robust estimator should possess a low $\lambda^{*}$.
\begin{mycor}\label{coro:lss}
$\lambda^{*}$ of Cauchy and Gaussian MLE estimators are bounded. $\lambda^{*}$ of Laplace MLE estimator is unbounded.
\end{mycor}
From Corollary \ref{coro:lss}, we can see Laplace estimator is very sensitive to local shifting around zero.

Cauchy MLE estimator has both bounded \textit{gross-error sensitivity} and bounded \textit{local-shift sensitivity}. Therefore, it is robust to gross noise and local shifting around zero.
\subsection{Optimization}
\begin{algorithm}[t]
\caption{Projected gradient descent for Cauchy PCA. } \label{svp}
\begin{algorithmic}

\STATE {\bfseries Input:} $M,k,\gamma, \text{tolerance}\, \epsilon, \text{step size}\, \eta $
\STATE {\bfseries Initialize:} $t=0, L^{t}=M$
\REPEAT
\STATE $G_{ij}^{t}=\frac{2(M_{ij}^{t}-L_{ij}^{t})}{\gamma^{2}+(M_{ij}^{t}-L_{ij}^{t})^{2}}$
\STATE $L^{t+1}\gets L^{t}-\eta G^{t}$
\STATE Compute top $k$ singular values and vectors of $L^{t+1}$: $U_{k},\Sigma_{k},V_{k}$
\STATE $L^{t+1}\gets U_{k}\Sigma_{k}V_{k}^{\mathsf{T}}$
\STATE $t\gets t+1$
\UNTIL $\|G\|_{F} <\epsilon$
\end{algorithmic}
\end{algorithm}
Gaining insights from \citet{meka2009guaranteed}, we adopt a projected gradient descent method to solve the problem defined in Eq.(\ref{eq:cpca}). It is an iterative approach where each iteration consists of gradient update and projection operation. Algorithm \ref{svp} outlines the optimization method. Low rank matrix $L$ to be estimated is initialized to the observation measurements $M$. At each iteration, we firstly compute the gradient matrix $G$, then use $G$ to update $L$. This is the ordinary gradient descent step. Then we project the newly obtained $L$ to the feasible set $\text{rank}(L)\leq k$. The projection is done by computing the largest $k$ singular values and singular vectors of $L$: $U_{k},\Sigma_{k},V_{k}$, then reconstructing $L$: $L=U_{k}\Sigma_{k}V_{k}^{\mathsf{T}}$. Note that in the projection phase, we only need to compute the top $k$ singular values and their corresponding singular vectors, which can be done efficiently by Lanczos SVD algorithm \citep{larsen1998lanczos}. Note that the optimization problem is not convex and may suffer local optimal. It would be helpful to run the algorithm multiple times with different random initializations.

\section{Experiments}
In this section, we first corroborate the ability of Cauchy PCA to recover matrices from various noise patterns on simulated data, then demonstrate its usage in a real application: face recognition with corruption.

\subsection{Simulation}
\label{sec:simul}
To evaluate the robustness of Cauchy PCA to various noise patterns, we generate low rank matrices, then corrupt them with noise of diverse amounts and magnitudes and try to recover them. We consider $n\times 2n$ matrices with $n=500,1000,2000$. Similar to \cite{candes2009robust}, we generate a rank $r=0.05\times n$ matrix $L_{0}=XY$ where $X\in R^{n\times r}$,$Y\in R^{r \times 2n}$ are matrices whose entries are independently sampled from uniform distribution $U(-1,1)$. 
To corrupt $L_{0}$, we randomly choose $\rho \times n \times 2n$ entries and each entry is independently added noise sampled uniformly from $[-m,m]$. We call $\rho\in [0,1]$ corruption rate and $m$ as noise magnitude. 
Each matrix is corrupted by 33 noise patterns by varying corruption rate $\rho=0,0.1,..,1$ and noise magnitude $m=0.1,1,10$. 
We compare Cauchy PCA with Gaussian PCA, Laplace PCA \citep{candes2009robust}, and multivariate Student-t PCA (MV-Student-t PCA) \citep{khan2004robust}. 
The rank constraint parameter $k$ in Gaussian PCA and Cauchy PCA and the dimension $d$ of latent vectors in MV-Student-t PCA are set to the intrinsic rank $r$ of the matrix to be recovered. 
Throughout the experiments, we set the scale parameter $\gamma$ of Cauchy PCA to 0.1. 
For Laplace PCA, we tune the trade-off parameter $\lambda$ and choose the largest one under which the estimated low rank matrix is of rank $r$.
We use $\|L_{0}-L\|_{F}/\| L_{0}\|_{F}$ to measure recovery error, where $L$ is the estimated low rank matrix and $L_{0}$ is the true low rank matrix.


\begin{figure}[t]
\begin{center}
   \includegraphics[width=0.7\linewidth]{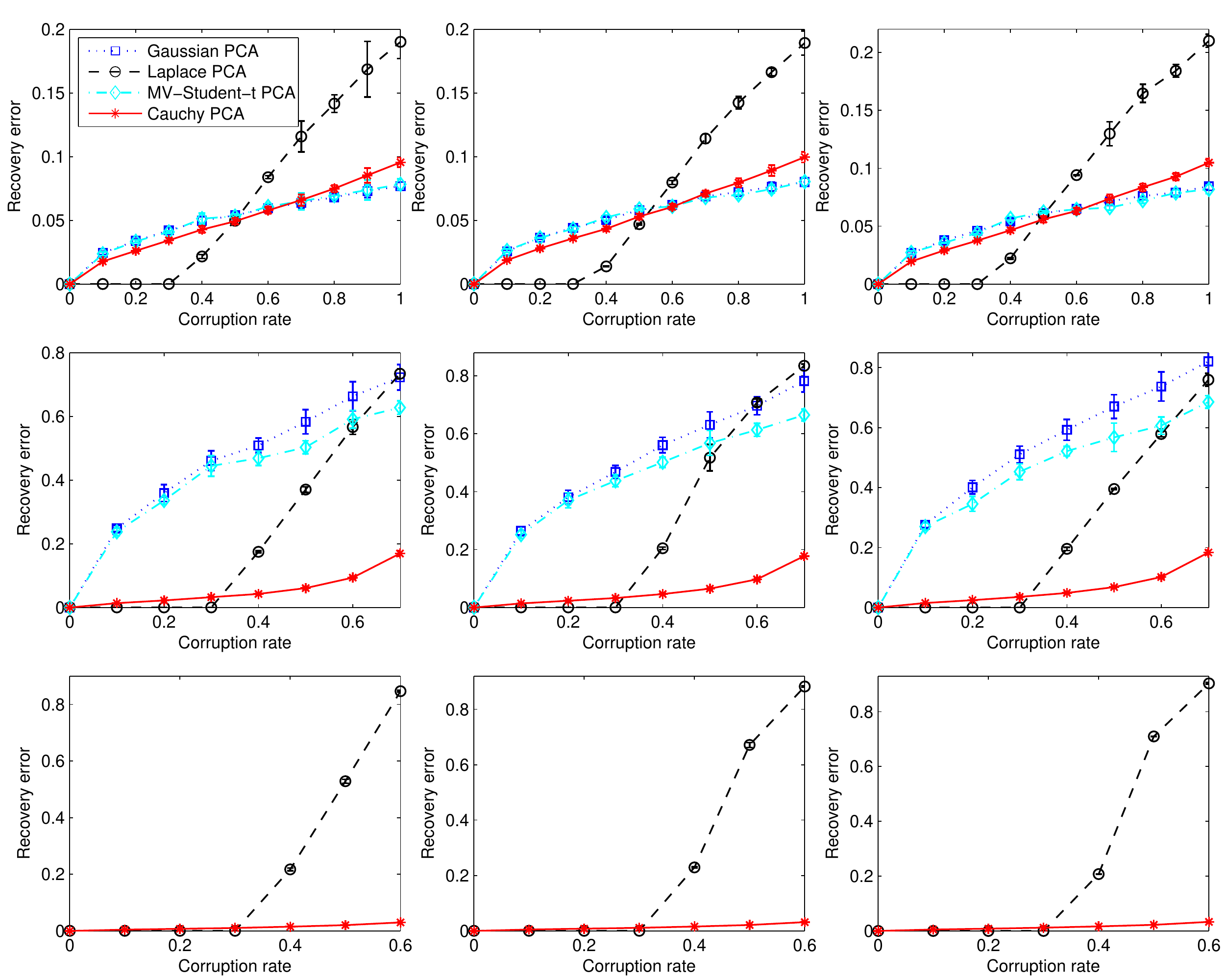}
\end{center}
   \caption{Matrix recovery under various noise patterns. Matrices in the first row are corrupted with noise magnitude $m=0.1$. In the second and third row, $m=1$ and $10$ respectively. Matrices in the first column is of size $n=500$. $n=1000, 2000$ in the second and third column respectively.}
\label{fig:simulation}
\vspace{-5mm}
\end{figure}

Figure \ref{fig:simulation} summarizes matrix recovery results of four PCA methods. Each subfigure shows recovery error $\|L_{0}-L\|_{F}/\| L_{0}\|_{F}$ versus corruption rate $\rho$. Noise magnitude $m=0.1,1,10$ in the first, second, third row respectively. Matrix size $n=500,1000,2000$ in the first, second, third column respectively. In the second row and third row, errors at $\rho>0.7$ and $\rho>0.6$ are not displayed since they are unreasonably high (greater than 1). In the third row, errors of Gaussian PCA and MV-Student-t PCA
at all corruption rate are not displayed for the same reason. As can be seen from the figure, Gaussian PCA quickly fails as noise becomes large.
When $m=10$, errors of Gaussian PCA are greater than 1 at all corruption rates greater than zero. 
In all cases, Laplace PCA works very well when noise is sparse but fails rapidly when noise becomes dense. When corruption rate $\rho$ is below 0.3, Laplace PCA can perfectly recover the low rank matrix. However, once $\rho$ exceeds 0.3, errors of Laplace PCA increase sharply. This corroborates our claim and analysis that Laplace PCA is only suitable for sparse noise. Cauchy PCA shows great robustness under all kinds of noise conditions. When noise is small (shown in the first row), Cauchy PCA has comparable performance with Gaussian PCA. When noise is large (shown in the second and third row), errors of Cauchy PCA are consistently small at all corruption rate. The performance of Cauchy PCA is significantly superior to the other two when noise is large and dense. 
For instance, when $m=10$, $\rho=0.6$, $n=1000$, the average error of Cauchy PCA is only 0.032 while Laplace PCA and Gaussian PCA suffer errors of 0.882 and 9.049.
The performance of MV-Student-t PCA is very similar to Gaussian PCA.
For small noise (the first row), MV-Student-t PCA is nearly the same as Gaussian PCA.
For large noise (the second and third row), MV-Student-t PCA works better than Gaussian PCA, but much worse than Laplace PCA and Cauchy PCA. We conjecture the reason is that in EM procedure of MV-Student-t PCA, each data instance is actually modeled using a Gaussian distribution with instance-specific variance, thereby, the final result is close to Gaussian PCA.

\subsection{Face Recognition With Corruption}\label{sec:face_recog}

In this section, we investigate the problem of face recognition where face images are contaminated by severe noise \citep{wright2009robust}.
The same as \cite{wright2009robust}, we randomly corrupt a percentage of pixels and evaluate the robustness of each PCA method in recognizing corrupted faces.
For face recognition, we adopt eigenface \citep{turk1991eigenfaces} methodology. Given the noisy training data, we use each PCA method to recover the low rank matrix, then obtain the basis from the low rank matrix. All training and testing images are projected into the low dimensional subspace spanned by the learned basis. For each testing face, recognition is performed by finding the nearest training face in the subspace and assigning the identity of the nearest training face to the testing face. We measure the recognition accuracy under varying corruption rate $\rho$. Recognition rate is defined as the ratio between the number of correctly recognized faces and the total number of test faces.
%
%
%
\begin{figure}[t]
\begin{center}
   \includegraphics[width=0.8\linewidth]{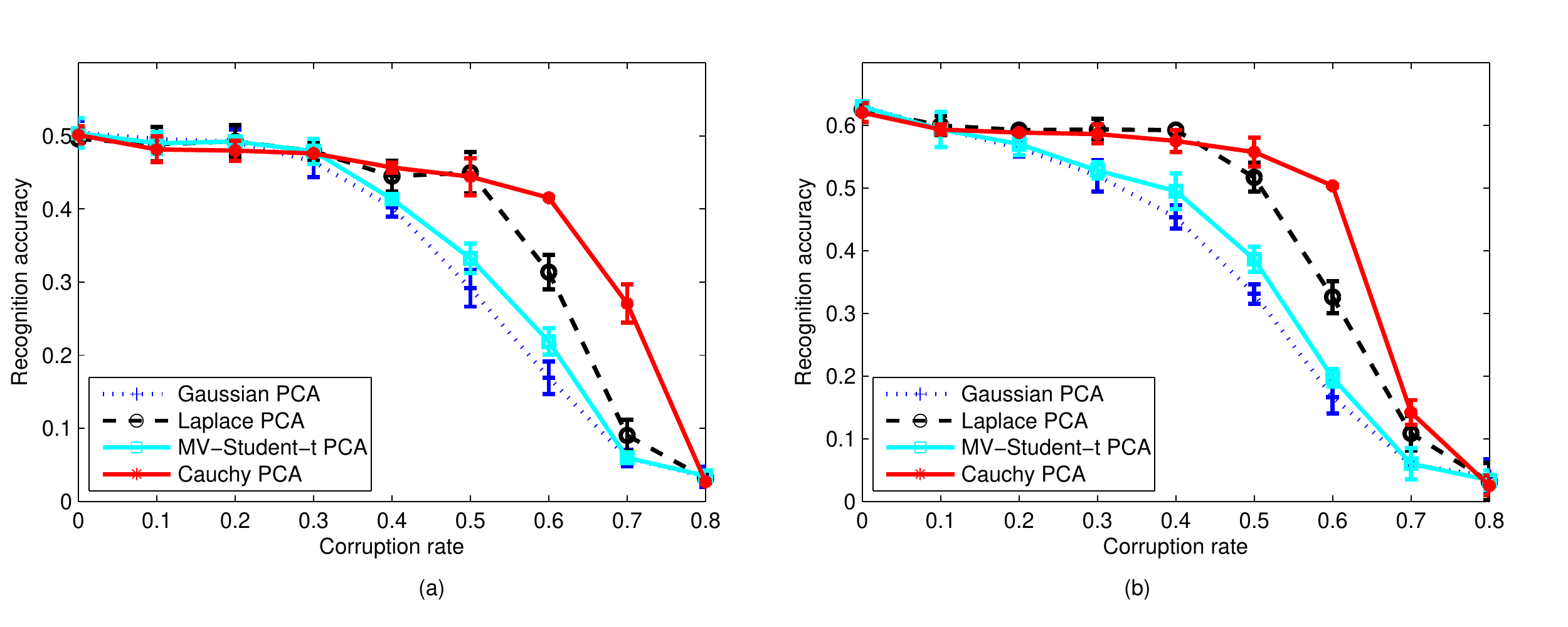}
\end{center}
   \caption{Face recognition accuracy under varying corruption rate. (a) rank constraint $k=30$. (b) rank constraint $k=60$.}
\label{fig:face_recog_accu}
\vspace{-5mm}
\end{figure}
We use the Extended Yale B dataset \citep{lee2005acquiring} consisting of 2414 frontal-face images of 38 individuals. For each individual, we randomly choose half images for training and the other half for testing. All images are resized to $96 \times 84$. 
Following \cite{wright2009robust}, we corrupt pixels by randomly replacing their values with integers sampled uniformly from $[0,255]$.
For each $\rho \in \{0,0.1,...,0.8\}$, we take 5 replications where for each image, $\rho\times 96\times 84$ pixels are corrupted. All corrupted images are normalized to have zero mean and unit standard deviation. We perform two experiments under different rank constraint settings. In the first experiment, we set the rank constraint $k$ of Gaussian PCA and Cauchy PCA and dimension $d$ of latent vectors in MV-Student-t to 30, and tune the tradeoff parameter of Laplace PCA to make sure the recovered low rank matrix is of rank 30. In the second experiment, rank constraint is set to 60.

Figure \ref{fig:face_recog_accu} shows the recognition accuracy under different corruption rate $\rho$. It can be seen that Cauchy PCA is more robust to dense large noise than Gaussian and Laplace PCA. For $k=30$ (Figure \ref{fig:face_recog_accu}(a)), Gaussian PCA quickly fails when $\rho$ exceeds 0.3. Cauchy and Laplace PCA have comparably stable performance when $\rho$ is below 0.5. At $\rho=0.6$, the accuracy of Laplace PCA has a sharp drop while Cauchy PCA remains stable. At $\rho=0.7$, the accuracy of Laplace PCA drops to 0.09 while Cauchy PCA achieves 0.27. Similar results can also be observed for $k=60$ (Figure \ref{fig:face_recog_accu}(b)). The recognition accuracy of MV-Student-t is better than Gaussian PCA and is worse than Laplace PCA and Cauchy PCA, which is consistent with the matrix recovery results reported in Section \ref{sec:simul}.
%
%
%

\begin{figure}[t]
\begin{center}
   \includegraphics[width=0.5\linewidth]{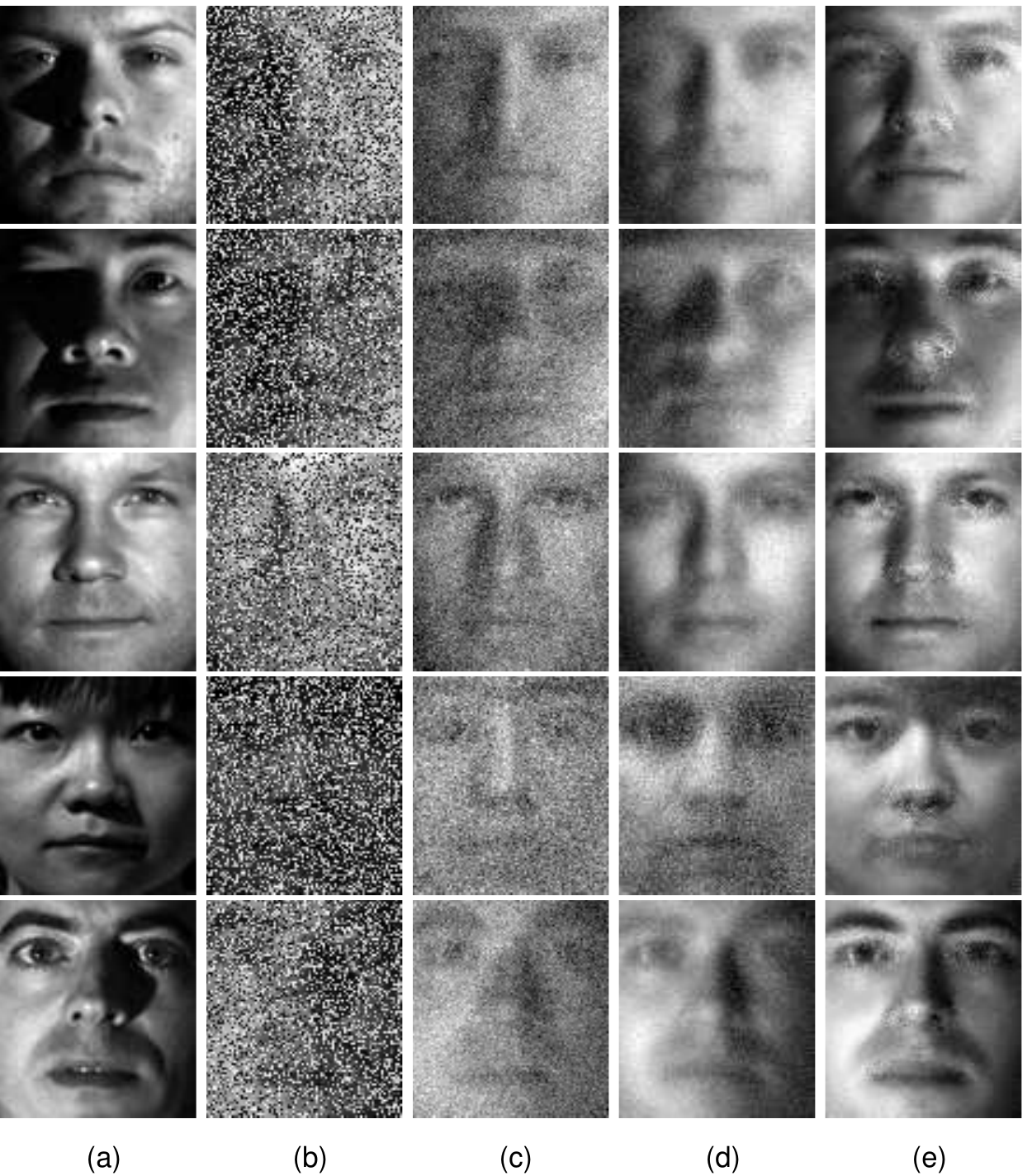}
\end{center}
   \caption{Face reconstruction results of three PCA methods (corruption rate $\rho=0.6$, rank constraint $k=30$). (a) Original face images. (b) Noise corrupted faces. (c) Reconstructed faces by Gaussian PCA. (d) Reconstructed faces by Laplace PCA. (e) Reconstructed faces by Cauchy PCA.}
\label{fig:face_recons}
\label{fig:face_recons}
\end{figure}

Figure \ref{fig:face_recons} shows face reconstruction results for $\rho=0.6$, $k=30$. Original face images (Figure \ref{fig:face_recons}(a)) are heavily corrupted by noise (Figure \ref{fig:face_recons}(b)). Reconstructed faces (Figure \ref{fig:face_recons}(c)) by Gaussian PCA are in severe contamination. Laplace PCA gets better results (Figure \ref{fig:face_recons}(d)), but the reconstructed images are still hard to recognize. Some reconstructions even change the original appearance. For example, reconstruction for the girl in the fourth row is completely wrong. In contrast, as shown in Figure \ref{fig:face_recons}(e), Cauchy PCA can successfully remove most noise and restore the original appearance. Reconstructed faces by Cauchy PCA are much easier to recognize.
The reconstruction results of MV-Student-t PCA are very close to those of Gaussian PCA. We do not show them to avoid skewing Figure \ref{fig:face_recons}.

\section{Conclusions}
We propose Cauchy principal component analysis, which is robust to various noise patterns. 
For large dense noise, Cauchy PCA significantly outperforms Gaussian PCA and Laplace PCA.
For small noise, Cauchy PCA has comparable performance with Gaussian PCA. For large noise, Cauchy PCA possesses comparable robustness with Laplace PCA.
Experiments on simulated data and real world applications corroborate our intuitive and theoretical analysis of the robustness of our method.
In future, we will seek further theoretical explanations and find more efficient solvers to scale Cauchy PCA to large datasets.

\bibliography{cpca}

\begin{thebibliography}{22}
\providecommand{\natexlab}[1]{#1}
\providecommand{\url}[1]{\texttt{#1}}
\expandafter\ifx\csname urlstyle\endcsname\relax
  \providecommand{\doi}[1]{doi: #1}\else
  \providecommand{\doi}{doi: \begingroup \urlstyle{rm}\Url}\fi

\bibitem[Archambeau et~al.(2006)Archambeau, Delannay, and
  Verleysen]{archambeau2006robust}
Archambeau, C{\'e}dric, Delannay, Nicolas, and Verleysen, Michel.
\newblock Robust probabilistic projections.
\newblock In \emph{ICML}, pp.\  33--40. ACM, 2006.

\bibitem[Brubaker(2009)]{brubaker2009robust}
Brubaker, S.C.
\newblock Robust pca and clustering in noisy mixtures.
\newblock In \emph{Proceedings of the twentieth Annual ACM-SIAM Symposium on
  Discrete Algorithms}, pp.\  1078--1087. Society for Industrial and Applied
  Mathematics, 2009.

\bibitem[Candes et~al.(2009)Candes, Li, Ma, and Wright]{candes2009robust}
Candes, E.J., Li, X., Ma, Y., and Wright, J.
\newblock Robust principal component analysis?
\newblock \emph{Arxiv preprint ArXiv:0912.3599}, 2009.

\bibitem[De~La~Torre \& Black(2003)De~La~Torre and Black]{de2003framework}
De~La~Torre, F. and Black, M.J.
\newblock A framework for robust subspace learning.
\newblock \emph{International Journal of Computer Vision}, 54\penalty0
  (1):\penalty0 117--142, 2003.

\bibitem[Ding et~al.(2006)Ding, Zhou, He, and Zha]{ding2006r}
Ding, C., Zhou, D., He, X., and Zha, H.
\newblock R 1-pca: rotational invariant l 1-norm principal component analysis
  for robust subspace factorization.
\newblock In \emph{ICML}, pp.\  281--288. ACM, 2006.

\bibitem[Eriksson \& Van Den~Hengel(2010)Eriksson and Van
  Den~Hengel]{eriksson2010efficient}
Eriksson, A. and Van Den~Hengel, A.
\newblock Efficient computation of robust low-rank matrix approximations in the
  presence of missing data using the l1 norm.
\newblock In \emph{CVPR 2010}, pp.\  771--778. IEEE, 2010.

\bibitem[Ganesh et~al.(2010)Ganesh, Wright, Li, Candes, and
  Ma]{ganesh2010dense}
Ganesh, A., Wright, J., Li, X., Candes, E.J., and Ma, Y.
\newblock Dense error correction for low-rank matrices via principal component
  pursuit.
\newblock In \emph{Information Theory Proceedings (ISIT), 2010 IEEE
  International Symposium on}, pp.\  1513--1517. IEEE, 2010.

\bibitem[Hampel(1974)]{hampel1974influence}
Hampel, F.R.
\newblock The influence curve and its role in robust estimation.
\newblock \emph{Journal of the American Statistical Association}, 69\penalty0
  (346):\penalty0 383--393, 1974.

\bibitem[Hampel et~al.(2011)Hampel, Ronchetti, Rousseeuw, and
  Stahel]{hampel2011robust}
Hampel, F.R., Ronchetti, E.M., Rousseeuw, P.J., and Stahel, W.A.
\newblock \emph{Robust statistics: the approach based on influence functions},
  volume 114.
\newblock Wiley, 2011.

\bibitem[Ke \& Kanade(2005)Ke and Kanade]{ke2005robust}
Ke, Q. and Kanade, T.
\newblock Robust l1 norm factorization in the presence of outliers and missing
  data by alternative convex programming.
\newblock In \emph{CVPR 2005}, volume~1, pp.\  739--746. IEEE, 2005.

\bibitem[Khan \& Dellaert(2004)Khan and Dellaert]{khan2004robust}
Khan, Zia and Dellaert, Frank.
\newblock Robust generative subspace modeling: The subspace t distribution.
\newblock 2004.

\bibitem[Larsen(1998)]{larsen1998lanczos}
Larsen, R.M.
\newblock Lanczos bidiagonalization with partial reorthogonalization.
\newblock 1998.

\bibitem[Lee et~al.(2005)Lee, Ho, and Kriegman]{lee2005acquiring}
Lee, K.C., Ho, J., and Kriegman, D.J.
\newblock Acquiring linear subspaces for face recognition under variable
  lighting.
\newblock \emph{Pattern Analysis and Machine Intelligence, IEEE Transactions
  on}, 27\penalty0 (5):\penalty0 684--698, 2005.

\bibitem[Meka et~al.(2009)Meka, Jain, and Dhillon]{meka2009guaranteed}
Meka, R., Jain, P., and Dhillon, I.S.
\newblock Guaranteed rank minimization via singular value projection.
\newblock \emph{Arxiv preprint arXiv:0909.5457}, 2009.

\bibitem[Salakhutdinov \& Mnih(2008)Salakhutdinov and
  Mnih]{salakhutdinov2008probabilistic}
Salakhutdinov, R. and Mnih, A.
\newblock Probabilistic matrix factorization.
\newblock \emph{Advances in neural information processing systems},
  20:\penalty0 1257--1264, 2008.

\bibitem[Tibshirani(1996)]{tibshirani1996regression}
Tibshirani, R.
\newblock Regression shrinkage and selection via the lasso.
\newblock \emph{Journal of the Royal Statistical Society. Series B
  (Methodological)}, pp.\  267--288, 1996.

\bibitem[Tipping \& Bishop(1999)Tipping and Bishop]{tipping1999probabilistic}
Tipping, M.E. and Bishop, C.M.
\newblock Probabilistic principal component analysis.
\newblock \emph{Journal of the Royal Statistical Society: Series B (Statistical
  Methodology)}, 61\penalty0 (3):\penalty0 611--622, 1999.

\bibitem[Tomasi \& Kanade(1992)Tomasi and Kanade]{tomasi1992shape}
Tomasi, C. and Kanade, T.
\newblock Shape and motion from image streams under orthography: a
  factorization method.
\newblock \emph{International Journal of Computer Vision}, 9\penalty0
  (2):\penalty0 137--154, 1992.

\bibitem[Turk \& Pentland(1991)Turk and Pentland]{turk1991eigenfaces}
Turk, M. and Pentland, A.
\newblock Eigenfaces for recognition.
\newblock \emph{Journal of cognitive neuroscience}, 3\penalty0 (1):\penalty0
  71--86, 1991.

\bibitem[Wright et~al.(2009)Wright, Yang, Ganesh, Sastry, and
  Ma]{wright2009robust}
Wright, John, Yang, Allen~Y, Ganesh, Arvind, Sastry, S~Shankar, and Ma, Yi.
\newblock Robust face recognition via sparse representation.
\newblock \emph{Pattern Analysis and Machine Intelligence, IEEE Transactions
  on}, 31\penalty0 (2):\penalty0 210--227, 2009.

\bibitem[Xu et~al.(2010)Xu, Caramanis, and Sanghavi]{xu2010robust}
Xu, H., Caramanis, C., and Sanghavi, S.
\newblock Robust pca via outlier pursuit.
\newblock \emph{Information Theory, IEEE Transactions on}, \penalty0
  (99):\penalty0 1--1, 2010.

\bibitem[Zhou et~al.(2010)Zhou, Li, Wright, Candes, and Ma]{zhou2010stable}
Zhou, Z., Li, X., Wright, J., Candes, E., and Ma, Y.
\newblock Stable principal component pursuit.
\newblock In \emph{Information Theory Proceedings (ISIT), 2010 IEEE
  International Symposium on}, pp.\  1518--1522. IEEE, 2010.

\end{thebibliography}
\bibliographystyle{iclr2015}

\end{document}